\documentclass[10pt,twocolumn,letterpaper]{article}

\usepackage[pagenumbers]{iccv} 

\usepackage{adjustbox}
\usepackage{csquotes}
\usepackage{epigraph}
\usepackage[most,skins,theorems]{tcolorbox}
\usepackage{colortbl}
\usepackage{float}
\usepackage{placeins}
\usepackage{stfloats}

\usepackage{listings}

\usepackage{multicol}
\usepackage{multirow}

\lstdefinestyle{pythonstyle}{
    language=Python,
    basicstyle=\ttfamily\footnotesize,
    keywordstyle=\color{blue},
    stringstyle=\color{green},
    commentstyle=\color{gray},
    morecomment=[l][\color{magenta}]{\#}
}

\definecolor{mygreen}{HTML}{3cb44b}
\definecolor{myred}{HTML}{800020}
\definecolor{myorange}{HTML}{FFBF00}

\definecolor{textcolor}{HTML}{049e88}
\definecolor{imagecolor}{HTML}{4c71c2}

\definecolor{mygray}{rgb}{0.89, 0.93, 0.85}
\definecolor{whitesmoke}{rgb}{0.96, 0.96, 0.96}
\definecolor{timberwolf}{rgb}{0.86, 0.84, 0.82}
\definecolor{darkgreen}{rgb}{0.0, 0.5, 0.0}
\definecolor{lightgray}{rgb}{0.9, 0.9, 0.9}
\definecolor{Mycolor1}{HTML}{BAD8F2}
\definecolor{Mycolor2}{HTML}{E0F0FA}

\newcolumntype{a}{>{\columncolor{mygray}}c}

\usepackage{tabu}
\usepackage{colortbl, array}
\usepackage{pgfplotstable}
\pgfplotsset{compat=1.8}

\definecolor{airforceblue}{rgb}{0.36, 0.54, 0.66}
\definecolor{battleshipgrey}{rgb}{0.52, 0.52, 0.51}

\newcommand{\topline}{ %
        \arrayrulecolor{rulecolor}\specialrule{0.1em}{\abovetopsep}{0pt}%
        \arrayrulecolor{tableheadcolor}\specialrule{\belowrulesep}{0pt}{0pt}%
        \arrayrulecolor{rulecolor}}
\newcommand{\midtopline}{ %
        \arrayrulecolor{tableheadcolor}\specialrule{\aboverulesep}{0pt}{0pt}%
        \arrayrulecolor{rulecolor}\specialrule{\lightrulewidth}{0pt}{0pt}%
        \arrayrulecolor{white}\specialrule{\belowrulesep}{0pt}{0pt}%
        \arrayrulecolor{rulecolor}}
\newcommand{\bottomline}{ %
        \arrayrulecolor{white}\specialrule{\aboverulesep}{0pt}{0pt}%
        \arrayrulecolor{rulecolor} %
        \specialrule{\heavyrulewidth}{0pt}{\belowbottomsep}}%

\pgfplotstableset{normal/.style ={%
        header=true,
        string type,
        column type=l,
        every odd row/.style={
            before row=
        },
        every head row/.style={
            before row={\topline\rowcolor{tableheadcolor}},
            after row={\midtopline}
        },
        every last row/.style={
            after row=\bottomline
        },
        col sep=&,
        row sep=\\
    }
}

\usepackage{graphicx,calc}
\newlength\myheight
\newlength\mydepth
\settototalheight\myheight{Xygp}
\tcbset{
  aibox/.style={
    width=\linewidth,
    top=8pt,
    bottom=4pt,
    colback=blue!6!white,
    colframe=black,
    colbacktitle=black,
    enhanced,
    center,
    attach boxed title to top left={yshift=-0.1in,xshift=0.15in},
    boxed title style={boxrule=0pt,colframe=white,},
  }
}
\newtcolorbox{AIbox}[2][]{aibox,title=#2,#1}
\definecolor{lightblue}{rgb}{0.22,0.45,0.70}

\newcommand{\name}{Fwd2Bot}

\definecolor{iccvblue}{rgb}{0.21,0.49,0.74}
\usepackage[pagebackref,breaklinks,colorlinks,allcolors=iccvblue]{hyperref}

\def\paperID{} 
\def\confName{ICCV}
\def\confYear{2025}

\title{Fwd2Bot: LVLM Visual Token Compression with Double Forward Bottleneck}

\author{%
  Adrian Bulat*$^{1,2}$ \quad Yassine Ouali*$^1$ \quad \quad Georgios Tzimiropoulos$^{1,3}$\\
$^1$Samsung AI Cambridge \quad $^2$Technical University of Iasi \quad $^3$Queen Mary University of London
}
\begin{document}
\maketitle
\begin{abstract}

In this work, we aim to compress the vision tokens of a Large Vision Language Model (LVLM) into a representation that is simultaneously suitable for (a) generative and (b) discriminative tasks, (c) is nearly lossless, and (d) is storage-efficient. We propose a novel compression approach, called Fwd2Bot, that uses the LVLM itself to compress the visual information in a task-agnostic manner. At the core of Fwd2bot there exists a ``double-forward pass'' training strategy, whereby, during the first forward pass, the LLM (of the LVLM) creates a bottleneck by condensing the visual information into a small number of summary tokens. Then, using the same LLM, the second forward pass processes the language instruction(s) alongside the summary tokens, used as a direct replacement for the image ones. The training signal is provided by two losses: an autoregressive one applied after the second pass that provides a direct optimization objective for compression, and a contrastive loss, applied after the first pass, that further boosts the representation strength, especially for discriminative tasks. The training is further enhanced by stage-specific adapters. We accompany the proposed method by an in-depth ablation study. Overall, Fwd2Bot results in highly-informative compressed representations suitable for both generative and discriminative tasks. For generative tasks, we offer a 2$\times$ higher compression rate without compromising the generative capabilities, setting a new state-of-the-art result. For discriminative tasks, we set a new state-of-the-art on image retrieval and compositionality.
\end{abstract}    
\let\thefootnote\relax\footnotetext{* Equal contribution. Listing order is random.}
\section{Introduction}
\label{sec:intro}

LVLMs (\ie LMMs or Visual LLMs) are LLMs~\cite{dubey2024llama,jiang2023mistral} that, in addition to text, are capable of integrating and processing visual information as input context~\cite{li2024llava}. They have been deployed for many tasks and use cases requiring multi-modal (\ie vision-language) understanding and reasoning like image captioning, visual question answering, and multi-modal chatbots. They are typically constructed by stitching and then fine-tuning together a pre-trained vision encoder (e.g. CLIP~\cite{radford2021learning}) and an LLM on conversational/instructional data, with the LLM jointly processing the visual and language tokens thereafter~\cite{li2024llava}. 

A key characteristic that distinguishes LVLMs from LLMs is that, for many tasks, the language instructions are short, and the input sequence length is dominated by the visual tokens. Hence, in this work, our aim is to answer the following question: \textit{``How can we compress LVLM's visual tokens into a short sequence of summary tokens in order to enable efficient deployment with minimal accuracy degradation?''} We refer to this problem as \textit{Token Compression}.

Several works have been recently proposed aiming to address the relevant problem of (what we call) \textit{Token Reduction}~\cite{li2024inference,shang2024llava,cai2024matryoshka,hu2024matryoshka}. Although Token Compression and Token Reduction have similar goals (\ie efficient deployment), they operate under different settings and trade-offs. Token Reduction assumes an on-the-fly setting whereby during inference, given an input image and the text instruction, the number of visual tokens must be effectively pruned (\ie reduced) before they are fed to the LLM. Token Compression distinguishes between offline and online processing. During offline processing, an image is compressed in a small number of visual summary tokens in a query/instruction agnostic manner which are stored for further processing. During inference (\ie online processing), the (saved) summary tokens can be directly provided along with the text instruction to the LLM (of the LVLM) for efficient processing of both generative and discriminative tasks. Although Token Reduction methods can also be run in offline-online mode, they do not utilize this distinction to improve compression efficiency, nor can they be used for discrimination.   

Going beyond Token Reduction methods, in this work, we set up a training procedure that specifically capitalizes on two-staged processing (\ie offline and online) by trading one-off offline processing with significantly superior accuracy gains (for the same number of summary tokens used) during online inference. Our main methodological contribution is that an excellent way to perform Token Compression is to use the LVLM itself to compress the visual tokens based on a ``double-forward pass'' strategy. Specifically, our method, called \name,  performs a first forward pass whereby the trainable summary tokens, the image tokens, and the tokens of the prompt ``Summarize the image in a few words.'' are all fed to the LVLM. This step allows the summary tokens to interact with the image and the prompt tokens by passing through all the layers of the LVLM. Then, we create a bottleneck by performing a second forward pass whereby only the summary tokens and the language instruction are passed through the LLM for optimization with next-token prediction loss. This step allows the summary tokens to be optimized directly for the task at hand. 

\begin{figure*}[ht!]
     \centering
     \begin{subfigure}[b]{0.3\textwidth}
         \centering
         \includegraphics[width=\textwidth]{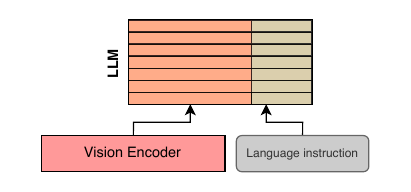}
         \vspace{0.3cm}
         \caption{LLaVA (baseline)~\cite{liu2024improved}: all vision tokens are processed for every query.}
         \label{fig:y equals x}
     \end{subfigure}
     \hfill
     \begin{subfigure}[b]{0.3\textwidth}
         \centering
         \includegraphics[width=\textwidth]{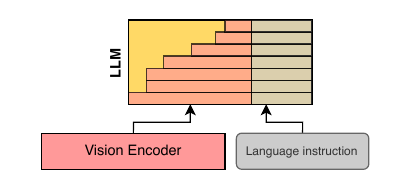}
         \caption{Adaptive token processing (\eg~\cite{chen2024image}): the vision tokens are pruned dynamically within each layer of the LLM. It requires all vision tokens (hence, storage is inefficient).}
         \label{fig:paradigm_adaptive}
     \end{subfigure}
     \hfill
     \begin{subfigure}[b]{0.3\textwidth}
         \centering
         \includegraphics[width=\textwidth]{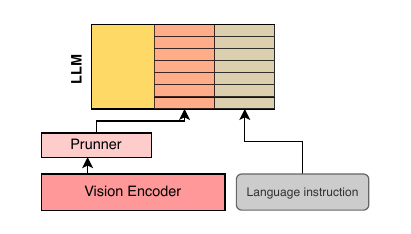}
         \caption{Direct vision token pruning (\eg~\cite{shang2024llava}): a separate module is learned separately or jointly with the vision encoder.}
         \label{fig:paradigm_direct}
     \end{subfigure}
     
     \begin{subfigure}[b]{0.4\textwidth}
         \centering
         \includegraphics[width=0.8\textwidth]{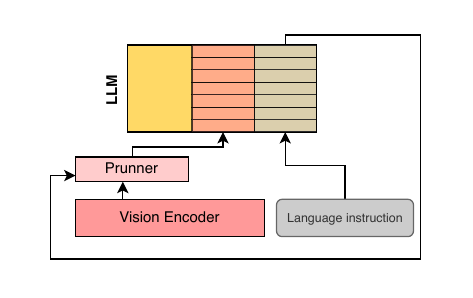}
         \caption{Query-dependent token compression (\eg~\cite{li2024inference}): the compression depends on the LLM features of each query.}
         \label{fig:paradigm_dependant}
     \end{subfigure}
     \quad
     \begin{subfigure}[b]{0.5\textwidth}
         \centering
         \includegraphics[width=\textwidth]{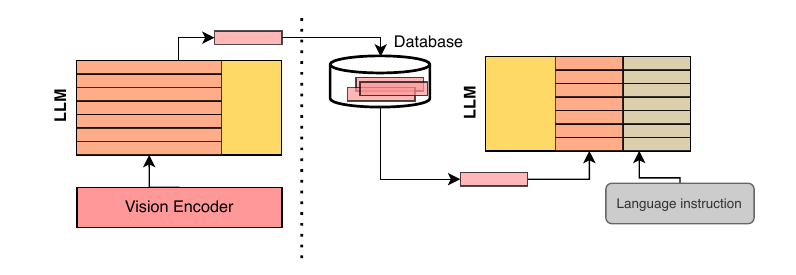}
         \caption{Our setting: the LVLM itself produces the compressed representations. Trades-off offline processing with superior compression performance.}
         \label{fig:paradigm_ours}
     \end{subfigure}
     \vspace{-0.2cm}
        \caption{Different methods/paradigms for reducing/compressing the visual tokens in LVLMs.}
        \label{fig:token_compression_paradigms}
        \vspace*{-0.4cm}
\end{figure*}

Going even further beyond previous works, a second methodological contribution of ours is that the summary tokens are optimized not only for autoregressive generation but also for discrimination (\ie image-text retrieval). For this purpose, we introduce a contrastive loss for optimizing the summary tokens. Importantly, we show that the contrastive loss is beneficial for enhancing the accuracy of the generative tasks, too.

Overall, \textbf{we make the following contributions}:
\begin{itemize}
\item 
We introduce Fwd2Bot, a new two-staged offline-online Token Compression method whereby highly condensed visual summary tokens are produced by the LVLM itself. For this purpose, we set up a training procedure whereby two forward passes through the LLM of the LVLM are utilized, creating a visual information bottleneck in-between that allows the training of the summary tokens for the task in hand. 
\item 
We show that the learned summary tokens produced by our method can be used not only for auto-regressive generation but also for image-text discrimination (\ie retrieval). For this purpose, we introduce a second loss for contrastive optimization. Importantly, we show that this loss further enhances the generative abilities of the model.
\item 
For generative tasks, our method achieves a 2x higher compression rate without compromising
the generative capabilities, significantly outperforming previous works. For discriminative tasks, we set a new state-of-the-art result for text-image retrieval and compositionality.
\end{itemize}

\section{Related work}

\subsection{Token Reduction in LVLMs} 

The current prevalent LVLM architecture consists of a pretrained LLM, a pretrained vision encoder, and a projector head that maps the vision tokens from the output space of the vision encoder to the input space of the LLM~\cite{liu2024visual,liu2024improved,wu2023visual,zhu2023minigpt,wang2023cogvlm,bai2023qwen,li2024mini,chu2024mobilevlm}. Despite their remarkable accuracy, a limitation of such approaches is their computational cost. In large part, this is a consequence of the LLM having to process a large number of vision tokens (\eg 576 in~\cite{liu2024improved}). To alleviate this, recently, a series of works focus on reducing their number ~\cite{li2024inference,shang2024llava,yu2024balancing,cai2024matryoshka,hu2024matryoshka} (see Fig.~\ref{fig:token_compression_paradigms} for a summary of concepts).

PruMerge~\cite{shang2024llava} reduces the number of vision tokens by discarding and/or merging the tokens produced by the vision encoder. The approach itself is training-free and relies on the similarities between the spatial local tokens and the global (CLS) one. In contrast, ~\cite{li2024tokenpacker} uses a learnable attentive module that combines and consolidates multi-level feature maps into a smaller set. \cite{cai2024matryoshka,hu2024matryoshka} learn nested representations using either a set of convolutions aimed to capture multi-resolution representation~\cite{cai2024matryoshka,chu2024mobilevlm} or a transformer layer~\cite{hu2024matryoshka}. ~\cite{wang2023cogvlm,chen2024image} dynamically vary the number of vision tokens within the LLM on a layer-by-layer basis using a rule-based system that takes into consideration the correlation between the tokens themselves or the tokens and the instruction.
Building on top of~\cite{li2024tokenpacker}, QueCC~\cite{li2024inference} conditions the token selection process on the user query, first processed by the LLM to produce a feature embedding before being used for conditioning of the compression module. 

Our method differs from the above works in several ways. Firstly, it is the only method designed and optimized to benefit from a two-staged process where an offline step is used to produce highly condensed summary tokens in a task-agnostic manner, which can be then efficiently used during inference. Secondly, it is the only method that sets up a training procedure that uses the whole LVLM itself to compress the visual tokens using a ``double-forward pass''
training strategy. Thirdly, it is the only method that learns summary representations for both generative and discriminative tasks. Finally, in terms of results, our method matches and surpasses the state-of-the-art of~\cite{li2024inference} without having to recompute the vision tokens for every new query.

\subsection{Discriminative LVLMs}

Very recently, a series of works~\cite{jiang2024e5,jiang2024vlm2vec,huang2024llm2clip} have explored the possibility of converting LVLMs into discriminative models, or of pairing LLMs with CLIP vision encoders. For example, ~\cite{huang2024llm2clip} directly aligns a pretrained LLM with a pretrained CLIP vision encoder. E5-V~\cite{jiang2024e5} through text-only contrastive training converts a generative LVLM into a discriminative one, while ~\cite{jiang2024vlm2vec} expands it to multi-modal retrieval. One major limitation of these approaches is the loss of generative abilities post-adaptation.  In this work, we address this very issue, creating a unified model that excels both at generative and discriminative tasks, surpassing recently proposed LVLMs adaptations for image-text retrieval and compositionality. 


\section{Method}

\subsection{Preliminaries}\label{ssec:method-prelminaries}

We implement our approach on top of the LLaVA-1.5~\cite{liu2024improved} model, leaving all architectural components unchanged. The LLaVA model consists of a pretrained CLIP vision encoder $g(.)$, a projection matrix $\mathbf{W}$, and an LLM $f(.)$. The input image $\mathbf{X}_v$ is passed to CLIP to produce vision embeddings $\mathbf{H}_v= g(\mathbf{X}_v) \mathbf{W}$. The language embeddings $\mathbf{H}_q$ are obtained from the input language instruction $\mathbf{X}_q$. Finally, the concatenated vision and language embeddings are passed to the LLM to compute the answer (output) embeddings $\mathbf{H}_a = f(\mathbf{H}_v; \mathbf{H}_q)$ which is decoded to the corresponding answer (output) sequence  $\mathbf{X}_a$.

Although autoregressive in nature, recently, it was shown that the model can be run in discriminative mode producing vision-text embeddings for matching \textit{\`a la} CLIP~\cite{jiang2024e5}. Using the prompt $\mathbf{X}_p$ ``summarize the above image in one word'' (or similar),
the vision embedding is produced as $\mathbf{e}_v = \mathbf{H}_a[-1], \; \mathbf{H}_a = f(\mathbf{H_v}; \mathbf{H}_p)$ ($\mathbf{H}_p$ is the language embedding of $\mathbf{X}_p$). Analogously, and given a text query $\mathbf{X}_{query}$, the text embedding is constructed as $\mathbf{e}_t = \mathbf{H}_a[-1],\; \mathbf{H}_a =  f(\mathbf{H}_{query}, \mathbf{H}_p$) ($\mathbf{H}_{query}$ is the embedding of a $\mathbf{X}_{query}$). As in CLIP~\cite{radford2021learning}, the image-text similarity is computed as $s = \mathrm{cos\_sim} (\mathbf{e}_v, \mathbf{e}_t)$.

\begin{figure*}[!ht]
    \centering
    \includegraphics[width=0.85\linewidth]{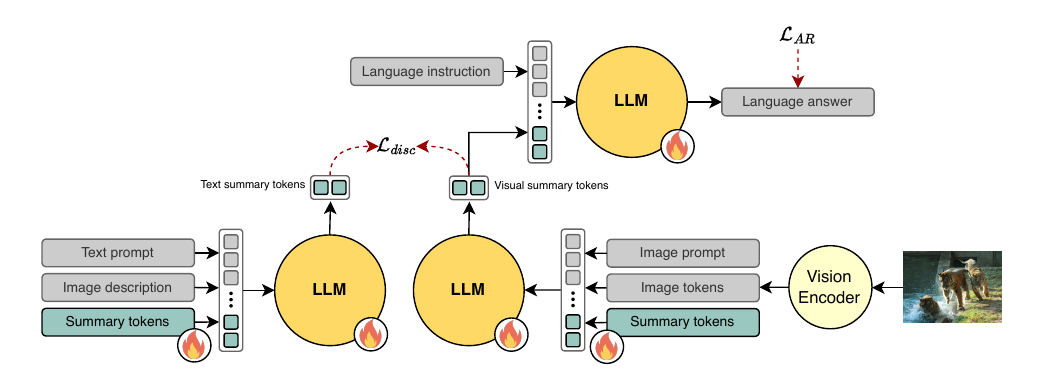}
    \caption{\textbf{Fwd2Bot training pipeline:} A first forward pass from the LLM (of the LVLM) creates a bottleneck by condensing the visual information into a small number of visual summary tokens. Then, using the same LLM (weights of depicted LLMs are shared), a second forward pass processes the language instruction(s) alongside the summary visual tokens for training with a next-token prediction loss $\mathcal{L}_{\mathrm{AR}}$ (see Sec.~\ref{ssec:method-double-fws}). Furthermore, a contrastive loss $\mathcal{L}_{\mathrm{disc}}$, applied after the first pass, is utilized to further boost the representation strength, especially for discriminative tasks (see Sec.~\ref{ssec:method-disctiminative-adaptation}). Components marked with \raisebox{-0.1cm}{\includegraphics[height=1.3em]{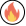}} are trainable. }
    \label{fig:enter-label}
\end{figure*}

\subsection{Double forward bottleneck algorithm}\label{ssec:method-double-fws}

The two axes influencing LVLM's efficiency are the model architecture/size and the input sequence length. In this work, we assume that the architecture is fixed. Hence, the inference cost is defined by the input sequence length, which turns out to be dominated by the length (number) $k$ of the vision embeddings $\mathbf{H_v} \in \mathbb{R}^{k \times d}$. Specifically, for a LLaVA model, $k=576$, which is significantly higher than the typical text query length and answer~\cite{li2024inference}. Hence, in this work, our goal is to derive a compressed visual token representation $\mathbf{H}^c_v\in \mathbb{R}^{k' \times d}$ where $k' \ll k$ without compromising the model's accuracy. Importantly, besides improving subsequent runs, a small sequence length also opens the path to offline pre-processing, whereby one can precompute, store, and re-use the compressed representation $\mathbf{H}^c_v$ without having to process the original image again.   

Several approaches for visual token compression have been proposed, including the introduction of new modules between the vision encoder and the LLM~\cite{hu2024matryoshka}, finetuning the vision encoder~\cite{cai2024matryoshka}, or compressing the vision tokens in an instruction/query dependent manner~\cite{li2024inference}. Departing from these approaches, we take a totally different path by proposing to leverage the LVLM itself (the LLM of the LVLM in particular) to self-compress the visual tokens. Our motivation for this is multifold. Firstly, LLMs have already excelled at text summarization~\cite{zhang2024benchmarking,liu2022revisiting}. Hence, we propose to utilize them for image summarization (\ie compression). However, as summarization with text, due to quantization (\ie tokenization), is inefficient with respect to the sequence length, we instead perform this summarization in a continuous latent space. 
Secondly, the LLM of the LVLM has already been utilized very recently to compute discriminant image-text embeddings ~\cite{jiang2024e5}. However, these embeddings cannot be used for generation. Hence, in this paper, based on a ``double-forward pass'' training strategy, we develop a simple, yet powerful, training bottleneck through which visual summary tokens at the output of the model are trained to highly compress visual information that can be used for both generation and discrimination. Overall, we call our method Fwd2Bot. See Fig.~\ref{fig:enter-label} for an overview.

More specifically, given an input image $\mathbf{X}_v$, we introduce  the summary tokens \ie learnable input embeddings $\mathbf{H}_r \in \mathbb{R}^{k' \times d}$ which evolve into the compressed vision embeddings $\mathbf{H}^c_v \in \mathbb{R}^{k' \times d}$ after \textit{a first forward pass from the LLM of the LVLM}:
\begin{equation}
    \label{Eq:fwd-pass-1}
    [;,;,\mathbf{H}^c_v] = f(\mathbf{H}_v; \mathbf{H}_p; \mathbf{H}_r),
\end{equation}
where $\mathbf{H}_p$ are the embeddings of the handcrafted prompt $\mathbf{X}_p$ ``Summarize the image in a few words.''. Clearly, during this pass, $\mathbf{H}_r$ interact with both $\mathbf{H}_v$ and $\mathbf{H}_p$. As the transformed embeddings $\mathbf{H}^c_v$ are query-agnostic, for subsequent language instructions/queries, the (same) LLM simply takes as input $\mathbf{H}^c_v$ instead of $\mathbf{H}_v$. 

To learn the compressed representation $\mathbf{H}^c_v$, during training, we perform \textit{a second forward pass from the LLM of the LVLM} where this time only $\mathbf{H}^c_v$ and the language instructions/embeddings are passed to the LLM. An autoregressive loss is applied at the output of the second forward pass: 
\begin{equation}
\mathcal{L}_{\mathrm{AR}} = - \sum^L_{i=1} \log \left( p_{\theta} (x_i|\mathbf{H}^c_v,\mathbf{X}_{q,<i},\mathbf{X}_{a,<i}) \right),
\vspace{-0.1cm}
\end{equation}
where $\theta$ represents the trainable parameters, $\mathbf{X}_{q,<i}$ and $\mathbf{X}_{a,<i}$ are the query/instruction and, respectively, answer tokens located before the current predicted token $x_i$. $\mathbf{H}_v$ is obtained from Eq.~\ref{Eq:fwd-pass-1}. Note that the weights of LLM are shared between the two forward passes. The flow of gradients through $\mathbf{H}^c_v$ results in a single model that can both compress and reason/generate answers by looking solely at the compressed tokens.

Intuitively, our algorithm can be also interpreted as a form of implicit chain-of-thought in the latent space~\cite{hao2024training}, with the LLM ``rephrasing'' the content of the vision sequence in a condensed manner for itself. Notably, while the input and output spaces of the LLM are not perfectly aligned, they are sufficiently close to resulting in good alignment of the compressed representations in just a few hundred iterations, making the whole training process efficient. That is the compressed representations simultaneously lie in the input and output space of the LVLM.

\subsection{Discriminative adaptation}\label{ssec:method-disctiminative-adaptation}

As noted above, an important conceptual difference between our approach and prior works in LVLM visual token reduction is that, in our work, the compressed representations lie simultaneously in both the input and output space of the LVLM. Unlike previous approaches, this enables us to directly leverage the compressed representations for CLIP-like discrimination in a zero-shot manner, as detailed in Sec.~\ref{ssec:method-prelminaries}. However, in this case, the discriminant performance is suboptimal as there is no explicit loss to encourage the separability of concepts.

To address this, and create a unified compressed representation suitable for both generative and discriminative tasks, we also propose to apply a contrastive loss over $\mathbf{H}^c_v$, at the output of the first forward pass. Not only does this loss enhance the discriminative properties of the compressed representation, but it can also improve the generative ability of the model thanks to learning a better underlying representation.

Given a dataset consisting of paired image-text samples,  the contrastive loss, for a given batch containing $B$ elements, is defined as:
\begin{equation}
    \mathcal{L}_{\mathrm{disc}} = \frac{1}{B}\sum_{k=1}^b (-\log \frac{\exp(s^{k,k}_v)}{\sum_{j} \exp(s^{k,j}_v)} - \log \frac{\exp(s^{k,k}_t)}{\sum_{j} \exp(s^{j,k}_t)}),
\end{equation}
where $s^{k,j}_v=\mathrm{cos\_sim}(\mathbf{e}_v^k, \mathbf{e}_t^j)$ computes the cosine similarity between the $k$-th image and the $j$-th caption. $\mathbf{e}_v = \frac{1}{k'} \sum \mathbf{H}^c_v$ and,  $\mathbf{e}_t = \frac{1}{k'} \sum  \mathbf{H}^c_t$, respectively. $\mathbf{H}^c_t$ is computed analogously to $\mathbf{H}^c_v$ (Eq.~\ref{Eq:fwd-pass-1}), except that it encodes textual data instead of visual, \ie $\mathbf{H}^c_t = f(\mathbf{H}_{query},\mathbf{H}_p,\mathbf{H}_r)$. $\mathbf{H}^c_t$ is only used as part of the discriminative loss.

\subsection{Overall training loss and data}\label{ssec:method-overall-loss}

The final model is trained using both losses, autoregressive and discriminative/contrastive:
\begin{equation}
    \mathcal{L}_{\mathrm{Total}} = \mathcal{L}_{\mathrm{AR}} + \mathcal{L}_{\mathrm{disc}}.
\end{equation}
At a given iteration, depending on the training data, the applicable losses are used. That is, for conversational data sampled from the LLaVA-665k dataset, we apply $\mathcal{L}_{\mathrm{AR}}$. For data sampled from CC3M, we apply  $\mathcal{L}_{\mathrm{disc}}$. If a conversational sample also has a caption associated with it, both losses are applied within the same iteration. For efficiency, the sampler will group together such cases. This also ensures that we have sufficiently large batches for contrastive training. The sampler aims for a 1:1 ratio between discriminative and autoregressive.

\subsection{Stage-specific adaptation}\label{ssec:method-task-aware-adaptation}

To enable efficient adaptation, we train our models using LoRA~\cite{hu2021lora} adapters which restrict the weight updates to a low-rank representation, $\Delta W = BA, \Delta W \in \mathbb{R}^{d\times m}, B\in\mathbb{R}^{d\times r}$ and $A\in\mathbb{R}^{r \times m}$, with $r << \min(d,m)$.

Although this works well, we use stage-specific adapters to further enhance the plasticity of the LVLM. We distinguish two stages that correspond to the two forward passes used during training: compression, which summarizes $\mathbf{H}_v$ into $\mathbf{H}^c_v$, and generation, which produces $\mathbf{X}_a$ given $\mathbf{H}^c_v$ and $\mathbf{X}_q$.
Depending on which operation we perform, compression or answer generation, different LoRA adapter weights $A$ and $B$ are used.

\section{Results} \label{sec:results}



\begin{table*}[!ht]
  \small
    \centering
      \caption{Comparison with various token reduction/compression methods on vision-language understanding tasks.}
    \label{tab:generative-evaluation}
    \vspace{-0.3cm}
    \begin{tabular}{|l|c|c|c|c|c|c|c|c|c|c|c|c|}
 \firsthline

         \rowcolor{airforceblue!20}
        \textbf{Method} & \# Tokens & GQA  & MMB  & MME    & POPE  & SQA  & TextVQA & VisWiz & VQAv2 \\
         \hline
LLAVA-1.5~\cite{liu2024improved} &  576  &  62.0 & 64.3 & 1510.7  & 85.9  & 66.8 & 58.2  &  50.0   & 78.5  \\
\hline
PruMerge~\cite{shang2024llava} & $\approx$32 & 57.2 & 60.9 & 1350.3  & 76.3  & 68.5 & \textbf{56.0} & 45.2   & 72.0  \\
TokenPacker~\cite{li2024tokenpacker}  & 36      & 59.6 & 62.8 & 1440.9 & 83.3 & \textbf{71.0} & 53.2  & 50.2 &  75.0  \\
Matryoshka Multi.~\cite{cai2024matryoshka}   & 36      & 60.3 & 64.8 & -       & 85.5  & -    & -      & 52.8 &    -  \\ 
 Matryoshka Query~\cite{hu2024matryoshka}   & 36      & 58.8 & 63.4 & 1416.3  & 81.9  & 66.8 & -     & 51.0 &   73.7  \\
 QueCC~\cite{li2024inference}       & 36      & 60.5 & 62.5 & 1442.0  & 84.5  & 70.6 & 53.3  & 50.1 & 75.8 \\
 \rowcolor{Mycolor1} \name~ (Ours)  &  32 & \textbf{61.6} & 64.6 & \textbf{1472.1} & \textbf{85.9} & 68.5 & 55.8 & \textbf{53.1} & \textbf{77.1}  \\

\hline
 TokenPacker~\cite{li2024tokenpacker}   &  16 & 58.9 &  62.7 & 1378.8 & 83.7 & 68.1 & 52.5  & 50.5 & 74.4 \\
 Matryoshka Query~\cite{hu2024matryoshka}  & 16  & 57.6 & 61.9 & 1408.5  & 80.8  & 67.5 & - & 49.8   & 71.1  \\
 QueCC~\cite{li2024inference}  & 16  & 59.0 & 62.2 & 1408.0  & 83.4  & \textbf{70.7} & 51.3 & 47.7  & 74.5 \\
 \rowcolor{Mycolor1} \name~ (Ours)  &  16  & \textbf{61.0} & \textbf{64.4} & \textbf{1470.0} & \textbf{85.6} & 67.7 & \textbf{54.2} & 49.8 & \textbf{76.5} \\
 \hline
  TokenPacker~\cite{li2024tokenpacker}    & 4  & 56.2 & 61.5 & 1347.6 & 81.7 & 68.5 & 49.2 & 45.7  & 70.5 \\
 Matryoshka Query~\cite{hu2024matryoshka}       & 4     & 53.0 & 56.5 & 1176.1  & 77.6 & 65.1 & -  & 49.4   & 64.1  \\
 QueCC~\cite{li2024inference}  & 4       & 56.5 & 62.1 & 1390.3 & 81.8& \textbf{68.6} & 48.7  & 45.0 & 70.6  \\
\rowcolor{Mycolor1} \name~ (Ours)  &  4 & \textbf{58.6} & \textbf{63.3} & \textbf{1403.0} & \textbf{84.3} & 67.7 & \textbf{52.5} & \textbf{51.6} & \textbf{74.5}  \\

        \lasthline
    \end{tabular}
    \vspace*{-0.2cm}
\end{table*}
\begin{table*}[!ht]
  \small
    \centering
      \caption{Comparison with various token reduction/compression methods on image captioning.}
    \label{tab:generative-captioning-evaluation}
    \vspace{-0.3cm}
     \resizebox{\textwidth}{!}{
    \begin{tabular}{|l|c|c|c|c|c|c|c|c|c|c|c|c|c|}
 \firsthline

         \rowcolor{airforceblue!20}
         &  & \multicolumn{4}{c|}{Flickr30K} & \multicolumn{4}{c|}{COCO} & \multicolumn{4}{c|}{nocaps}  \\
        \cline{3-14} 
        \rowcolor{airforceblue!20}
         \multirow{ -2}{*}{Method} &  \multirow{ -2}{*}{\# Tokens} & B@4 & CIDEr & MET. & ROUGE & B@4 & CIDEr & MET. & ROUGE  & B@4 & CIDEr & MET. & ROUGE\\
         \hline
LLAVA-1.5~\cite{liu2024improved} &  576  &  30.6 & 	81.2 & 	25.0 & 	53.4 & 32.9 & 	115.4 & 	27.7 & 	56.3 & 42.9 & 	105.3 & 	28.9 & 	59.8  \\
\hline
PruMerge~\cite{shang2024llava} & $\approx$32 &18.5 &	36.3 & 	15.7 & 40.2 & 18.5 & 66.3 & 	18.8 & 	44.9 & 25.9 & 	58.6 & 	20.0 & 	47.8 \\
Matryoshka Multi.~\cite{cai2024matryoshka}   & 36      & 25.4 	& 68.7 	& 24.1 	& 49.9 & 27.7 & 	102.2 & 	27.2 & 	53.3 & 36.8 & 	93.6 &	28.0 &	56.5\\ 
 Matryoshka Query~\cite{hu2024matryoshka}   & 36      & 26.4 	& 69.5 & 	23.1 &	50.0 & 28.0 & 	101.3 & 	26.2 & 	52.7 & 36.2 &	90.0 & 	26.8 &	55.8 \\
 \rowcolor{Mycolor1} \name~ (Ours)  &  32 & \textbf{30.0} & 	\textbf{78.9} & 	\textbf{25.2} &  	\textbf{52.9}  & \textbf{31.5}  &	\textbf{113.1} & 	\textbf{27.9} & 	\textbf{55.6} & \textbf{42.5} & 	\textbf{105.9} & 	\textbf{29.2} & \textbf{59.6} 	 \\

\hline
 Matryoshka Query~\cite{hu2024matryoshka}  & 16  & 24.8 & 	65.2 & 	22.7 &	49.0 & 27.6 & 	99.2 & 	26.0 & 	52.5 & 36.2 & 	90.0 & 	26.8 & 	55.8  \\
 \rowcolor{Mycolor1} \name~ (Ours)  &  16  & \textbf{29.0} & 	\textbf{78.2} & 	\textbf{25.3} & 	\textbf{52.7} & 	\textbf{31.0} & 	\textbf{112.0} & 	\textbf{27.9} & 	\textbf{55.4} & 	\textbf{42.0} & 	\textbf{104.7} & 	\textbf{29.3} & 	\textbf{59.5}\\
 \hline
 Matryoshka Query~\cite{hu2024matryoshka}       & 4     & 20.1 & 	47.5 & 	19.8  &	44.5 & 23.2 & 	81.0 & 	23.0 & 	48.6 &  	28.4 & 	63.2 & 	21.1 & 	49.5 \\
\rowcolor{Mycolor1} \name~ (Ours)  &  4 & \textbf{28.4} & 	\textbf{74.5} & 	\textbf{24.8} & 	\textbf{51.8} & \textbf{31.1} & 	\textbf{111.4} & 	\textbf{27.9} & 	\textbf{55.4} & \textbf{41.1} & 	\textbf{103.4} & 	\textbf{29.0} & 	\textbf{59.1} \\

        \lasthline
    \end{tabular}
    }
    \vspace*{-0.0cm}
\end{table*}

\begin{table*}[!ht]
  \small
    \centering
      \caption{Zero-shot text-image retrieval accuracy on Flickr30K, COCO and nocaps.}
    \label{tab:zero-shot-retrieval}
    \vspace{-0.3cm}
     \resizebox{\textwidth}{!}{
    \begin{tabular}{|l|c|c|c|c|c|c|c|c|c|c|c|c|}
 \firsthline
 \rowcolor{airforceblue!20}
        & \multicolumn{6}{c|}{\textbf{image} retrieval} & \multicolumn{6}{c|}{\textbf{text} retrieval} \\
        \cline{2-7} \cline{8-13}
         \rowcolor{airforceblue!20}
        \textbf{Method} & \multicolumn{2}{c|}{Flickr30K} & \multicolumn{2}{c|}{COCO} & \multicolumn{2}{c|}{nocaps} & \multicolumn{2}{c|}{Flickr30K} & \multicolumn{2}{c|}{COCO} & \multicolumn{2}{c|}{nocaps}  \\
        \cline{2-3} \cline{4-5} \cline{6-7} \cline{8-9} \cline{10-11} \cline{12-13}
        \rowcolor{airforceblue!20}
        & R@1 & R@10 & R@1 & R@10 & R@1 & R@10 & R@1 & R@10  & R@1 & R@10 & R@1 & R@10\\
         \hline
         \rowcolor{battleshipgrey!20}
          \multicolumn{13}{|c|}{Contrastive approaches} \\
          \hline
        CLIP (\texttt{ViT-B})~\cite{radford2021learning} & 58.8 & 89.8 & 30.5  & 66.8 & 46.8 & 85.1 & 77.8 & 98.2 & 51.0 & 83.5 & 67.3 & 95.6  \\
        CLIP (\texttt{ViT-L})~\cite{radford2021learning} & 67.3 & 93.3 & 37.0 & 71.5 & 48.6 & 85.7 & 87.2 & 99.4 & 58.1 & 87.8 & 70.0 & 96.2\\
        BLIP (\texttt{ViT-L})~\cite{li2022blip} & 70.0  & 95.2 & 48.4 & 83.2 & 62.3 & 93.4 & 75.5 & 97.7 & 63.5  & 92.5  & 72.1 & 97.7\\
        BLIP2 (\texttt{ViT-L})~\cite{li2023blip} & 74.5  & 97.2 & 50.0 & 86.1 & 63.0 & 93.8 & 86.1 & 99.4 & 63.0  & 93.1 & 74.4 & 98.3 \\
        OpenCLIP (\texttt{ViT-G/14})~\cite{schuhmann2022laion} & 77.8 & 96.9 & 48.8 & 81.5 & 63.7 & 93.2 & 91.5 & 99.6 & 66.3 & 91.8 & 81.0 & 98.7\\
        OpenCLIP (\texttt{ViT-BigG/14})~\cite{schuhmann2022laion} & 79.5 & 97.1 & 51.3 & 83.0 & 65.1 & 93.5 & 92.9 & 97.1 & 67.3 & 92.6 & 82.3 & 98.8 \\
        EVA-02-CLIP (\texttt{ViT-E/14+})~\cite{sun2023eva} & 78.8 & 96.8 & 51.1 & 82.7 & 64.5 & 92.9 & 93.9 & 99.8 & 68.8 & 92.8 & 83.0 & 98.9 \\
        EVA-CLIP~\cite{sun2024eva} & 80.3 & 97.2 & 52.0 & 82.9 & 65.3 & 93.2 & 94.5 & 99.7 & 70.1 & 93.1 & 83.5 & 98.6 \\
         \hline
         \rowcolor{battleshipgrey!20}
          \multicolumn{13}{|c|}{LVLM-based approaches} \\
          \hline
        LLaVA-1.5-7B~\cite{liu2024improved} & 59.6 & 89.3 & 34.4 & 69.6 & 46.9 & 83.3 & 65.6 & 92.3 & 35.6 & 70.5 & 52.1 & 88.1 \\
        E5-V (\texttt{LLaVA-1.5-7B})~\cite{jiang2024e5} & 76.7 & 96.9 & 48.2 & 82.1 & 62.0 & 93.0 & 86.6 & 99.0 & 57.4 & 88.4 & 71.9 & 97.0 \\
        VLM2Vec (\texttt{Mistral-7B})~\cite{jiang2024vlm2vec} & 80.1 & 97.3 & 52.0 & 85.6 & 65.9 & 94.5 & 90.3 & 99.6 & 68.2 & 93.2 & 79.2 & 98.5 \\
        \rowcolor{Mycolor1} \name~(Ours) (\texttt{LLaVA-1.5-7B}) & \textbf{83.8} & \textbf{98.5} & \textbf{59.0} & \textbf{88.6} & \textbf{72.3} & \textbf{96.5}  & 94.3 & \textbf{99.9} & \textbf{72.9} & \textbf{94.4} & \textbf{85.7} & \textbf{99.5}\\
        \lasthline
    \end{tabular}
    }
\end{table*}

\subsection{Generative benchmarks}

Following~\cite{liu2024improved}, we evaluate our approach on a diverse collection of datasets, mainly: GQA~\cite{hudson2019gqa}, MMB~\cite{liu2024mmbench}, MME~\cite{liang2024survey}, POPE~\cite{li2023evaluating}, SQA~\cite{lu2022learn}, TextVQA~\cite{singh2019towards}, VisWiz~\cite{gurari2018vizwiz} and VQAv2~\cite{goyal2017making}. To ensure fairness, in all cases, we fully align the test-time settings and processing  with~\cite{liu2024improved}. 
In addition to this, we also evaluate our approach for captioning on MS-COCO~\cite{lin2014microsoft}, Flickr30k~\cite{young2014image} and NoCaps~\cite{agrawal2019nocaps}, comparing it to token reduction methods that have models openly available. See supplementary material for results on TextCaps~\cite{sidorov2020textcaps}.

When comparing our approach with the state-of-the-art token reduction methods for visual-language understanding, as the results from Tab.~\ref{tab:generative-evaluation} show, we set a new best result, outperforming prior works using 2.25$\times$ fewer tokens (16 vs 36). Our results for $32$ and even $16$ tokens nearly match the uncompressed LlaVA~\cite{liu2024improved} baseline.

Similarly, when evaluated for zero-shot captioning (Tab.~\ref{tab:generative-captioning-evaluation}), our approach matches LLaVA's accuracy, significantly outperforming prior methods. This suggests that the proposed approach encodes more information in its compressed tokens. We note that LLaVA saw some MS-COCO images during training; hence, the MS-COCO evaluation is not fully zero-shot for all methods listed.

\subsection{Discriminative benchmarks}

To measure the discriminative abilities of our model, we evaluate it on a diverse set of retrieval benchmarks:
Flicr30k~\cite{young2014image}, MS-COCO~\cite{lin2014microsoft}, NoCaps~\cite{agrawal2019nocaps} and SugarCrepe~\cite{hsieh2024sugarcrepe}. The last one measures the compositional capabilities of the model, an area where CLIP and CLIP-like models tend to under-perform.

To offer a full picture, our method is compared against a diverse set of contrastive models, varying in size (from 0.2B parameters for CLIP (ViT-B)~\cite{radford2021learning} to 8B for EVA-CLIP~\cite{sun2024eva}) and architecture (CLIP~\cite{radford2021learning}, BLIP~\cite{li2023blip} and LVLM-based~\cite{jiang2024e5}). Our approach matches and outperforms all of them.  In particular, for zero-shot image-to-text and text-to-image retrieval, as the results from Tab.~\ref{tab:zero-shot-retrieval} show, we match and outperform larger models, \ie EVA-CLIP (8B vs. 7.06B), despite using 3 orders of magnitude fewer samples for training (2,700M for EVA-CLIP vs. $\sim$3M for ours). A similar trend can be observed when evaluated for compositionality on SugarCreppe (Tab.~\ref{tab:sota_eval_sc}). An interesting observation about the latter is that models derived from LVLMs (\eg, E5-V and ours) demonstrate significantly stronger compositionality. This suggests that the LVLM run in discriminative mode inherits the strong vision-language understanding of the underlying generative model.

\begin{table*}[!htbp]
  \centering
  \small
  \caption{Comparison with state-of-the-art on the SugarCrepe compositionality benchmark.}\label{tab:sota_eval_sc}
  \vspace{-0.3cm}
  \resizebox{0.9\textwidth}{!}{
    \begin{tabular}{|l|c|c| c|c|c|c|c|c|c|} 
    \firsthline
     \rowcolor{airforceblue!20}
       & Params & \multicolumn{3}{c|}{ Replace}  &  \multicolumn{2}{c|}{ Swap} & \multicolumn{2}{c|}{ Add} \\ 
     \cline{3-5}\cline{6-7}\cline{8-9}
      \rowcolor{airforceblue!20}
    \multirow{ -2}{*}{Method} & (B)  & Object & Attribute & Relation & Object & Attribute & Object & Attribute   \\ 
         \hline
         \rowcolor{battleshipgrey!20}
          \multicolumn{9}{|c|}{Contrastive approaches} \\
          \hline
NegCLIP \cite{yuksekgonul2022and} & 0.15 & 92.7 & 85.9 & 76.5 & 75.2 & 75.4 & 88.8 & 82.8\\
CLIP (\texttt{ViT-B})~\cite{radford2021learning} & 0.15 & 90.9 & 80.1 & 69.2 & 61.4 & 64.0 & 77.2 & 68.8\\
   CLIP (\texttt{ViT-L})~\cite{radford2021learning} & 0.43 & 94.1 & 79.2 & 65.2 & 60.2 & 62.3 & 78.3 & 71.5\\
   BLIP (\texttt{ViT-L})~\cite{li2022blip} & 0.23 & 96.5 & 81.7 & 69.1 & 66.6 & 76.8 & 92.0 & 85.1  \\
   BLIP2 (\texttt{ViT-L})~\cite{li2023blip} & 1.17 & 97.6 & 81.7 & 77.8 & 62.1 & 65.5 & 92.4 & 87.4 \\
   OpenCLIP (\texttt{ViT-G/14})~\cite{schuhmann2022laion} & 1.37 & 95.8& 85.0& 72.4& 63.0& 71.2& 91.5& 82.1\\
   OpenCLIP (\texttt{ViT-BigG/14})~\cite{schuhmann2022laion} & 2.54 & 96.6& 87.9& 74.9& 62.5& 75.2& 92.2& 84.5\\
   EVA-02-CLIP (\texttt{ViT-E/14+})~\cite{sun2023eva} & 5.04 & 97.1 & 88.5 & 74.2 & 67.3 & 74.1 & 91.8 & 83.9 \\
   EVA-CLIP~\cite{sun2024eva} & 8.22 & 96.4 & 86.6 & 74.8 & 66.1 & 74.6 & 91.3 & 82.0 \\
         \hline
         \rowcolor{battleshipgrey!20}
          \multicolumn{9}{|c|}{LVLM-based approaches} \\
          \hline
LLaVA-1.5-7B~\cite{liu2024improved} & 7.06 & 88.0 & 81.6 & 76.1 & 60.9 & 58.8 & 67.0 & 62.4 \\
E5-V (\texttt{LLaVA-1.5-7B})~\cite{jiang2024e5} & 7.06 & 95.8 & 86.6 & 81.6 & 62.9 & 64.0 & 93.5 & 88.0 \\
VLM2Vec (\texttt{Mistral-7B})~\cite{jiang2024vlm2vec} & 7.3 & 97.2 & 89.0 & 81.7 & 62.9 & 72.5 & 94.7 & 88.6\\
\rowcolor{Mycolor1} \name~ (Ours) (\texttt{LLaVA-1.5-7B}) & 7.06 & \textbf{98.1} & \textbf{89.5} & \textbf{82.7} & \textbf{77.8} & \textbf{78.1} & \textbf{95.3} & \textbf{93.1} \\
\lasthline
\end{tabular}
}
\vspace*{-0.25cm}
\end{table*}

\section{Ablation studies \& analysis}

In this section, we ablate the main components of our approach. See supplementary material for more ablations.

\begin{table}[!htbp]
  \centering
  \small
  \caption{Effect of generative and discriminative losses for generation (MMB, MME, TextVQA) and retrieval (Flickr30K, MS-COCO).}\label{tab:ablation_loss}
  \vspace{-0.3cm}
  \resizebox{0.47\textwidth}{!}{
    \begin{tabular}{|l|c|c| c|c|c|c|c|c|c|} 
    \firsthline
     \rowcolor{airforceblue!20}
       &  &   &  & \multicolumn{2}{c|}{ Flickr30K} & \multicolumn{2}{c|}{ MS-COCO} \\ 
     \cline{5-6}\cline{7-8}
      \rowcolor{airforceblue!20}
    \multirow{ -2}{*}{Method} & \multirow{ -2}{*}{MMB}  & \multirow{ -2}{*}{MME} & \multirow{ -2}{*}{TextVQA} & T2I & I2T & T2I & I2T   \\ 
    \hline
Discriminative & 46.2 & 624.3 & 13.5 & 84.3 & 94.8 & 56.3 & 73.2  \\
Generative  & 64.1 & 1420.1 & 54.2 & 61.3 & 76.0 & 33.9 & 47.0   \\
Both  & 64.4 & 1470.0 & 54.2 & 83.8 & 94.4 & 56.8 & 70.2  \\
\lasthline
\end{tabular}
}
\vspace{-0.4cm}
\end{table}

\noindent \textbf{Impact of each loss function:} As detailed in Secs.~\ref{ssec:method-double-fws} and~\ref{ssec:method-disctiminative-adaptation}, our models are trained using two losses: one autoregressive, applied after the second forward pass, and one contrastive, applied over the compressed tokens, after the first pass. In Tab.~\ref{tab:ablation_loss}, we report results for a LLaVA model evaluated using 16 tokens on generative and discriminative tasks. Intuitively, training solely with the discriminative loss (1st row) results in degraded generative performance, as no alignment between the input and output space of the LLM is performed. Moreover, discriminative losses applied over short captions tend to focus on coarse details, missing out on more fine-grained details. Conversely, applying only the generative loss (2nd row) results in degraded retrieval abilities, as no loss explicitly encourages concept separation. We note that the longer the training scheduler is, the more pronounced these degradations are for the two cases.

Finally, combining the two losses (3rd row) results in the best performance across the board. Notice that the two losses are complementary when applied jointly and boost the model's accuracy on both sets of benchmarks.

\begin{table}[!htbp]
  \centering
  \small
  \caption{Single vs. stage-specific LoRA v.s full finetuning for generation (MMB, MME, TextVQA) and retrieval (Flickr30K, MS-COCO).}\label{tab:ablation_adapters}
  \vspace{-0.3cm}
  \resizebox{0.47\textwidth}{!}{
    \begin{tabular}{|l|c|c| c|c|c|c|c|c|c|} 
    \firsthline
     \rowcolor{airforceblue!20}
       &  &   &  & \multicolumn{2}{c|}{ Flickr30K} & \multicolumn{2}{c|}{ MS-COCO} \\ 
     \cline{5-6}\cline{7-8}
      \rowcolor{airforceblue!20}
    \multirow{ -2}{*}{Method} & \multirow{ -2}{*}{MMB}  & \multirow{ -2}{*}{MME} & \multirow{ -2}{*}{TextVQA} & T2I & I2T & T2I & I2T   \\ 
    \hline
Fine-tuning & 64.3 & 1413.1 & 52.9 & 83.1 &  94.0 & 56.2 & 70.4  \\
Single LoRA  & 64.3 & 1410.5 & 51.8  & 83.8 & 94.1 & 56.5 & 69.9   \\
Stage LoRA  & 64.4 & 1470.0 & 54.2 & 83.8 & 94.4 & 56.8 & 70.2  \\
\lasthline
\end{tabular}
}
\vspace{-0.1cm}
\end{table}

\begin{figure}[!ht]
    \centering
    \includegraphics[width=0.85\linewidth]{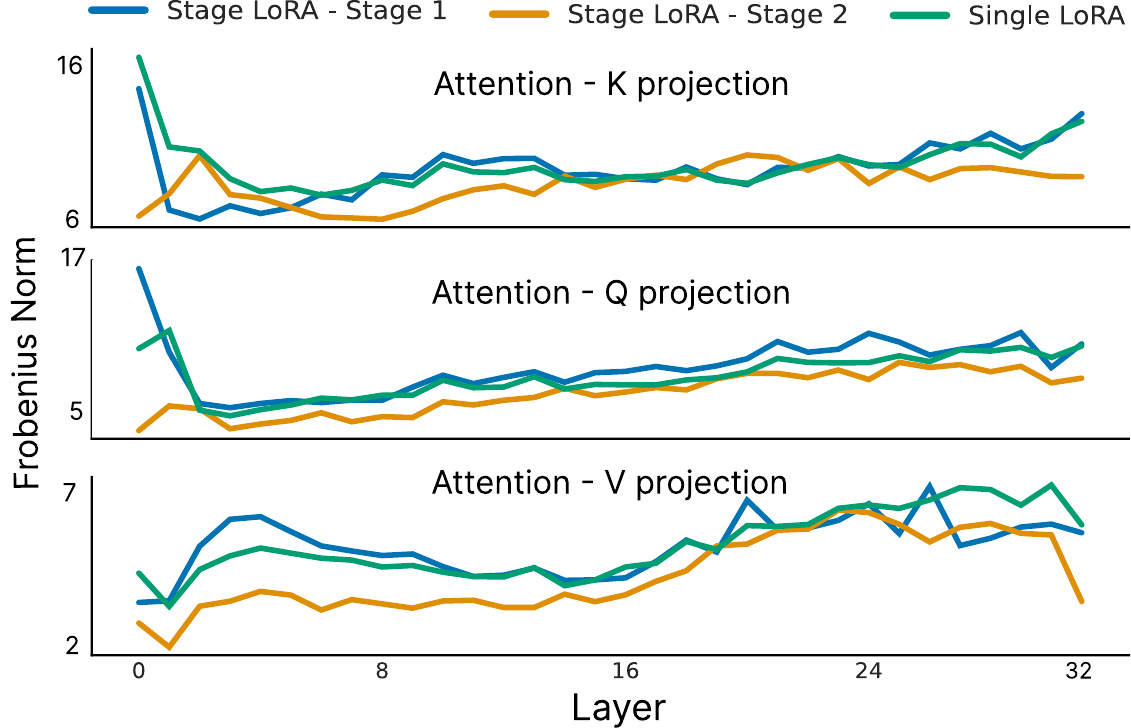}
    \vspace{-0.3cm}
    \caption{The norm of the learned LoRA weights adjustment $\Delta W = B A$ for a model trained with either a single LoRA or stage-specific LoRAs.}
    \label{fig:loras}
    \vspace{-0.4cm}
\end{figure}

\begin{figure*}[!ht]
    \centering
    \includegraphics[width=1.0\linewidth]{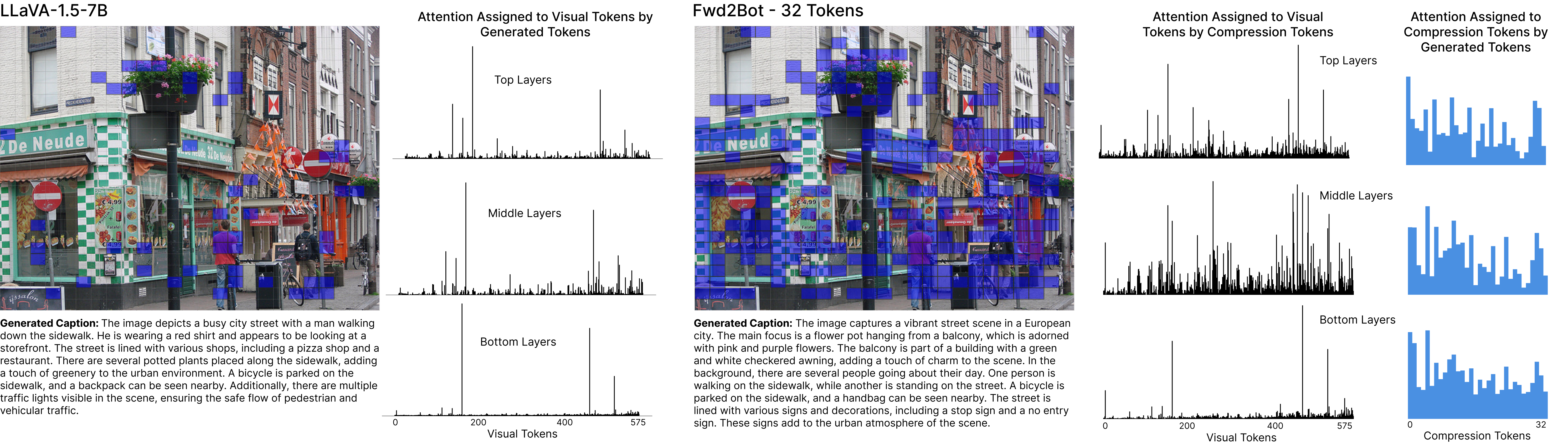}
    \vspace*{-0.7cm}
    \caption{Visualization of attention weights assigned to the 576 visual tokens and the 32 compressed tokens. On the left, we show the cumulative weights assigned to each visual token by the generated tokens for the baseline LLaVa-1.5-7B model. For Fwd2Bot, on the right, we first display the per-visual-token weights assigned by the summary tokens during the first forward pass to produce the summary tokens. We then show the weights assigned to the compressed tokens by the generated tokens during the second forward pass.}
    \label{fig:attention_weights}
\end{figure*}

\noindent \textbf{Single vs. stage-specific LoRA vs. full fine-tuning:} Herein, we compare the effect of training using (a) a single shared LoRA adapter, (b) stage-specific adapters, as proposed in Sec.~\ref{ssec:method-task-aware-adaptation}, and (c) full fine-tuning. We present the results of these choices in Tab.~\ref{tab:ablation_adapters}. The best results are obtained using the stage-specific adapters. The fine-tuning run suffers from overfitting to some extent, and its larger training cost makes optimization more difficult.

Fig.~\ref{fig:loras} further solidifies the need for stage-specific LoRAs, as the optimal representations required during compression (first forward pass) vs downstream inference (second forward pass) are different, especially for earlier layers.

\begin{figure}[!ht]
    \centering
    \includegraphics[width=1.0\linewidth]{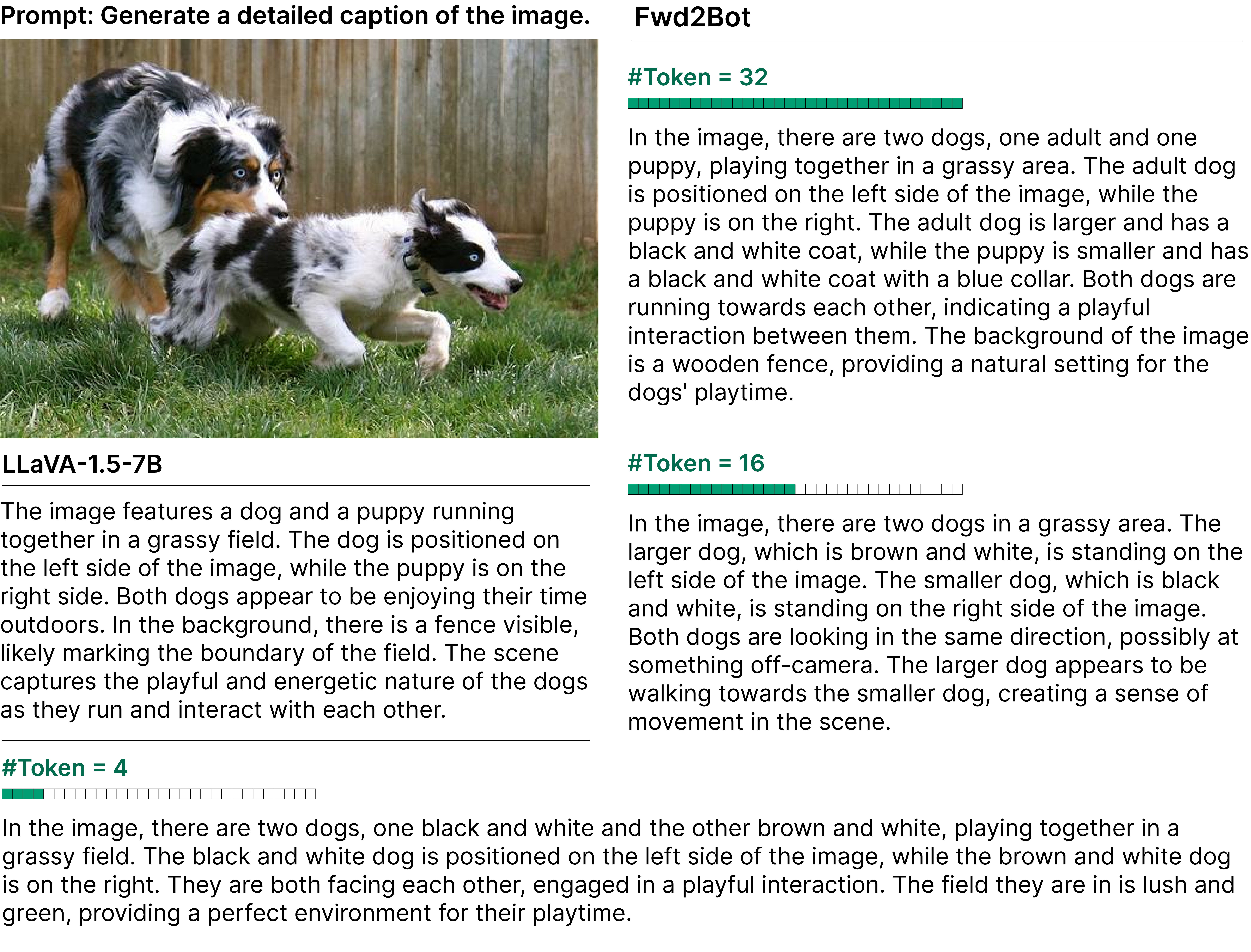}
    \vspace{-0.8cm}
    \caption{Captioning with variable number of summary tokens.}
    \label{fig:impact_of_num_tokens}
\end{figure}

\begin{figure}[!ht]
    \centering
    \includegraphics[width=0.75\linewidth]{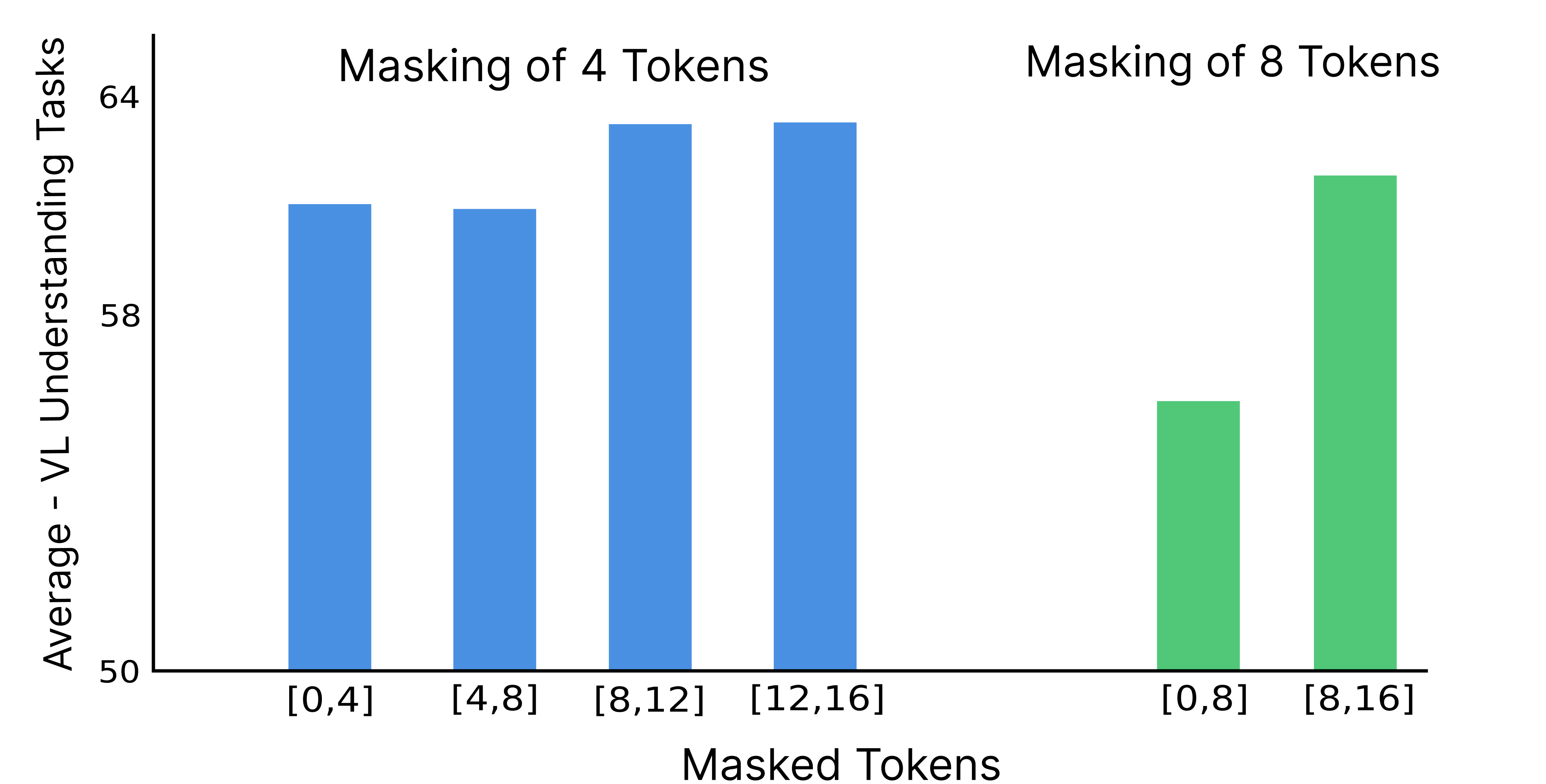}
    \vspace{-0.2cm}
    \caption{The relative importance of different subsets of visual tokens. We show the mean over the VL understanding tasks when masking specific subsets of the compressed visual tokens.}
    \label{fig:relative_importance_of_subset}
    \vspace{-0.3cm}
\end{figure}

\noindent \textbf{Double forward vs single forward:} To showcase the importance of our ``double-forward pass'' training strategy, we conducted the following experiment: instead of using the LLM itself to compress the vision summary tokens, we use the CLIP vision encoder only, effectively compressing them in the input space of the LLM. In this case, the loss is directly applied after the LLM, as in LLaVA, using a single forward pass. As shown in Tab.~\ref{tab:2stage-vs-1}, this baseline (1st row) vs ours (2nd row) performs significantly worse.

\begin{table}[!htbp]
  \centering
  \small
  \caption{Double vs single forward pass for generation (MMB, MME, TextVQA) and retrieval (Flickr30K, MS-COCO).}\label{tab:2stage-vs-1}
  \vspace{-0.3cm}
  \resizebox{0.47\textwidth}{!}{
    \begin{tabular}{|l|c|c| c|c|c|c|c|c|c|} 
    \firsthline
     \rowcolor{airforceblue!20}
       &  &   &  & \multicolumn{2}{c|}{ Flickr30K} & \multicolumn{2}{c|}{ MS-COCO} \\ 
     \cline{5-6}\cline{7-8}
      \rowcolor{airforceblue!20}
    \multirow{ -2}{*}{Method} & \multirow{ -2}{*}{MMB}  & \multirow{ -2}{*}{MME} & \multirow{ -2}{*}{TextVQA} & T2I & I2T & T2I & I2T   \\ 
    \hline
Single-forward stage & 61.3 & 1305.0 & 42.1 & 80.8 & 92.1 & 53.1 & 66.2   \\
Double-forward (Ours)  & 64.4 & 1470.0 & 54.2 & 83.8 & 94.4 & 56.8 & 70.2 \\
\lasthline
\end{tabular}
}
\end{table}

\noindent \textbf{What do the compressed tokens encode?} The compressed representation gradually encodes, from left to right, coarser to finer-grained concepts. This effect can be observed in Fig.~\ref{fig:impact_of_num_tokens}, where, as the number of tokens increases, the caption generated correctly captures more elements present in the photo, importantly reducing hallucinations. This effect is also corroborated in Fig.~\ref{fig:relative_importance_of_subset}. There, we mask-out different groups of (4 and 8) tokens, quantitatively measuring the impact of this: earlier tokens induce larger drops in performance (\eg masking the first 8 tokens reduces performance by 10\%). However, the performance does not drop to (near) 0, which suggests that there is also some degree of redundancy. 
The observed behavior can largely be attributed to the causal attention masking used by the LVLM, which encourages a directional information distribution.

\noindent \textbf{How does the model behavior change?} To shed light on the changes the model undergoes to act as a self-compressor, we analyze the attention patterns before and after our fine-tuning. The results of this visual analysis are presented in Fig.~\ref{fig:attention_weights}. Looking on the left side, we can observe that LLaVA exhibits a sparse attention pattern across all layers, particularly early on. In contrast, during self-compression, our model attends to all visually important parts of the image, having a significantly denser attention pattern at all layers. Intuitively, in the first case, as the model has access to all tokens, during generation, the model can peak back at the vision tokens as needed. In contrast, during compression, the LLM must ensure that all visually important details are stored in the compressed representation.
Finally, on the right-most part of the figure, we showcase the attention pattern between the generated tokens and the compressed representation obtained during text generation from compressed representations. Earlier and late-most tokens tend to have a higher attention weight.

\section{Conclusions}

In this work, we introduced \name, a novel LVLM visual token compression approach that uses the LVLM itself to compress the visual information in a task-agnostic manner, that is trained using a new ``double-forward pass'' training strategy. This results in a compressed visual representation that is simultaneously suitable for (a) generative and (b) discriminative tasks, (c) is nearly lossless, and (d) is storage-efficient. 
Performance-wise, for generative tasks, we offer a 2$\times$ higher compression rate without compromising the generative capabilities, setting a new state-of-the-art. For discriminative tasks, we also set a new state-of-the-art result on image retrieval and compositionality.
{
    \small
    \bibliographystyle{ieeenat_fullname}
    \bibliography{main}
}


\appendix

\paragraph{Results for larger models:} In the main manuscript, we conduct experiments using a LLaVA-1.5 (7B) model. Herein, we validate how our approach behaves when using a larger model, \ie a LLaVA-1.5 (13B). As the results from Table~\ref{tab:appendix-generative-evaluation} show, the proposed method nearly matches the full LLaVA model's accuracy using only 16 and even 4 tokens in this case, too.

\begin{table*}[!hb]
  \small
    \centering
      \caption{Token compression performance on vision-language understanding tasks using a LLaVA-1.5 13B model.}
    \label{tab:appendix-generative-evaluation}
    \begin{tabular}{|l|c|c|c|c|c|c|c|c|c|c|c|c|}
 \firsthline

         \rowcolor{airforceblue!20}
        \textbf{Method} & \# Tokens & GQA  & MMB  & MME    & POPE  & SQA  & TextVQA & VisWiz & VQAv2 \\
         \hline
LLAVA-1.5~\cite{liu2024improved} &  576  &  63.3 & 67.7 & 1531.0  & 86.2  & 71.6 & 61.3  &  53.6   & 80.0  \\
\hline
\name~ (Ours)  &  32 & 62.2 & 67.6 & 1465.1 & 85.3 & 72.5 & 59.6 & 54.0 & 78.7  \\
 \name~ (Ours)  &  16  & 61.8 & 67.3 & 1473.5 & 85.0 & 72.4 & 57.5 & 54.2 & 78.4 \\
\name~ (Ours)  &  4 & 59.9 & 66.4 & 1390.1 & 84.4 & 71.1 & 53.6 & 52.7 & 75.9  \\

        \lasthline
    \end{tabular}
    \vspace*{-0.2cm}
\end{table*}

\begin{figure*}[!hb]
    \centering
    \includegraphics[width=0.95\linewidth]{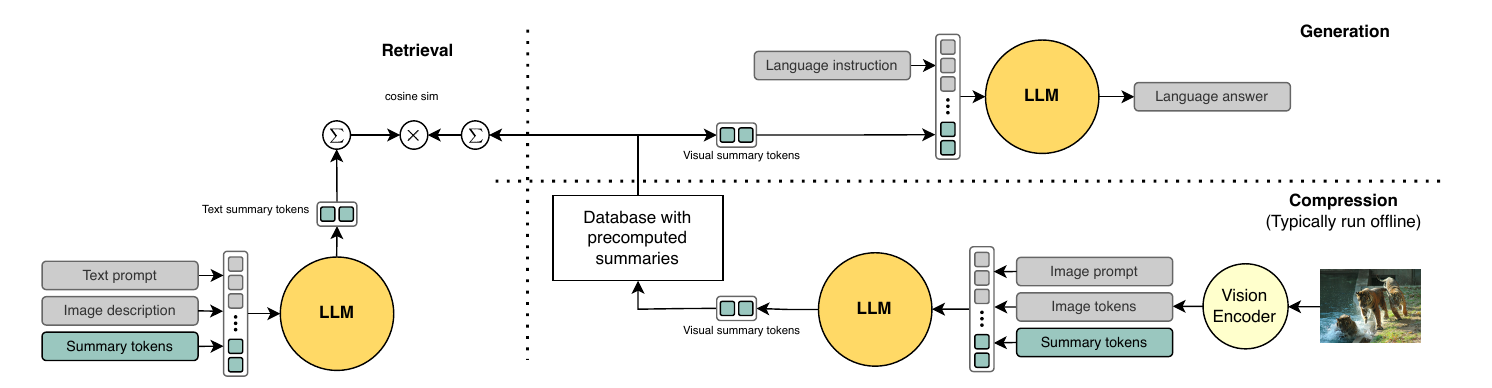}
    \caption{Test-time inference, depicting: compression (lower-right), generation (upper-right) and discrimination (left). Notice that in all cases we use the same LLM. The compressed embeddings are suitable for both set of tasks and are generally expected to be pre-computed offline.}
    \label{fig:appendix-inference}
\end{figure*}

\paragraph{Additional details regarding test-time inference:} 
In Fig.~\ref{fig:appendix-inference}, we depict the test-time inference flow for generative and discriminative tasks. The first step compresses the given image $\mathbf{X}_v$ into its compressed representation $\mathbf{H}^c_v$. This representation is then stored in a database. Note that while $\mathbf{H}^c_v$ can be computed on the fly, too, the scenario we are mostly interested in is pre-indexing, whereby the image representations are computed offline ahead of time.

Once stored, we can directly operate on this compressed representation for both generative and discriminative tasks. For generative tasks, the same LLM used for compression takes as input a user-provided instruction and the compressed image representation, producing an answer autoregressively (Fig.~\ref{fig:appendix-inference}, top-right corner).
For discriminative tasks, in order to measure the text-to-image similarity, \'a la CLIP, and again using the same LLM, we pass the user query (image description) to the LLM, producing a set of embeddings. We can measure the similarity between the given image description by taking the cosine similarity between the sum of the precomputed compressed vision tokens and the text embeddings newly produced by the LLM (Fig.~\ref{fig:appendix-inference}, left side).

\paragraph{Efficiency analysis:} Generally, for a transformer model~\cite{vaswani2017attention}, the total inference-time compute cost~\cite{kaplan2020scaling}, excluding nonlinearities, biases, normalizations, residuals and embeddings, which are negligible is equal to:  $\texttt{FLOPs} \approx \mathcal{O}( N \cdot T)$. In other words, the number of FLOPs depends on the number of parameters, $N$, and the sequence length, $T$. For a more detailed breakdown, see~\cite{kaplan2020scaling}.

Hence, for a LLaVA model, the total inference cost is approximatively: $\texttt{FLOPs}_{LLaVA} \approx \mathcal{O}( N_{v} \cdot V) + \mathcal{O}( N_{LLM} \cdot (V + Q + G)).$ $N_v$ and $N_{LLM}$ represents the number of parameters of the vision, and LLM, respectively. $V$ denotes the number of vision tokens (576 in LLaVA), $Q$ of query/instruction tokens, and $G$ of generated tokens.

For our approach, the initial cost to obtain the compressed representation $\mathbf{H}^c_v$ is equal to $\mathcal{O}( N_{v} \cdot V) + \mathcal{O}( N_{LLM} \cdot (V + K))$, with $K << V$ representing the number of compressed tokens. This process is aimed to be run offline, as a pre-indexing step, resulting in the storage of $2Kd$ bytes per image. We note that this representation is directly suitable for both generation and retrieval.
Once computed, the cost of subsequent queries is equal to: $\mathcal{O}( N_{LLM} \cdot (K + Q + G))$. In contrast, the current state-of-the-art approach, QueCC~\cite{li2024inference}, requires  $\mathcal{O}( N_{LLM} \cdot (K + 2Q + G))$ due to the dependency on the user query/instruction for token compression. Moreover, from a storage point of view, in~\cite{li2024inference}, all V (576) tokens must be stored or, alternatively, recomputed if an image from the database is queried again.

\begin{table*}[!ht]
  \small
    \centering
      \caption{Zero-shot text-image retrieval accuracy on Flickr30K, COCO and nocaps.}
    \label{tab:appendix-zero-shot-retrieval}
     \resizebox{\textwidth}{!}{
    \begin{tabular}{|l|c|c|c|c|c|c|c|c|c|c|c|c|c|}
 \firsthline
 \rowcolor{airforceblue!20}
        & & \multicolumn{6}{c|}{\textbf{image} retrieval} & \multicolumn{6}{c|}{\textbf{text} retrieval} \\
        \cline{3-8} \cline{9-14}
         \rowcolor{airforceblue!20}
        \textbf{Method} & \textbf{Tokens} & \multicolumn{2}{c|}{Flickr30K} & \multicolumn{2}{c|}{COCO} & \multicolumn{2}{c|}{nocaps} & \multicolumn{2}{c|}{Flickr30K} & \multicolumn{2}{c|}{COCO} & \multicolumn{2}{c|}{nocaps}  \\
        \cline{3-4} \cline{5-6} \cline{7-8} \cline{9-10} \cline{11-12} \cline{13-14}
        \rowcolor{airforceblue!20}
        & & R@1 & R@10 & R@1 & R@10 & R@1 & R@10 & R@1 & R@10  & R@1 & R@10 & R@1 & R@10\\
         \hline
        LLaVA-1.5-7B~\cite{liu2024improved} & 576 & 59.6 & 89.3 & 34.4 & 69.6 & 46.9 & 83.3 & 65.6 & 92.3 & 35.6 & 70.5 & 52.1 & 88.1 \\
        PruMerge~\cite{shang2024llava} & 18 & 34.7 & 67.9 & 18.4 & 47.9 & 25.8 & 62.7 & 38.3 & 74.3 & 19.8 & 49.9 & 28.2 & 65.2 \\
        Matryoshka Multi.~\cite{cai2024matryoshka} & 16 & 57.9 & 88.5 & 34.1 & 69.7 &45.5 & 83.2 & 63.8 & 91.7 & 36.4 & 72.5 & 48.0 & 86.2   \\
        Matryoshka Query~\cite{hu2024matryoshka} & 16 & 53.6 & 85.9 & 29.8 & 65.4 & 40.5 & 80.0 & 59.4 & 90.3 & 34.1 & 69.6 & 45.4 & 84.7\\
        \rowcolor{Mycolor1} \name~(Ours)  & 16 & \textbf{83.8} & \textbf{98.5} & \textbf{59.0} & \textbf{88.6} & \textbf{72.3} & \textbf{96.5}  & 94.3 & \textbf{99.9} & \textbf{72.9} & \textbf{94.4} & \textbf{85.7} & \textbf{99.5}\\
        \lasthline
    \end{tabular}
    }
    \vspace*{-0.2cm}
\end{table*}

\begin{table*}[!htbp]
  \centering
  \small
  \caption{Comparison on the SugarCrepe compositionality benchmark.}\label{tab:appendix-sota_eval_sc}
  \resizebox{0.9\textwidth}{!}{
    \begin{tabular}{|l|c|c| c|c|c|c|c|c|c|} 
    \firsthline
     \rowcolor{airforceblue!20}
       &  & \multicolumn{3}{c|}{ Replace}  &  \multicolumn{2}{c|}{ Swap} & \multicolumn{2}{c|}{ Add} \\ 
     \cline{3-5}\cline{6-7}\cline{8-9}
      \rowcolor{airforceblue!20}
    \multirow{ -2}{*}{Method} & \multirow{ -2}{*}{Tokens}  & Object & Attribute & Relation & Object & Attribute & Object & Attribute   \\ 
         \hline
LLaVA-1.5-7B~\cite{liu2024improved} & 576 & 88.0 & 81.6 & 76.1 & 60.9 & 58.8 & 67.0 & 62.4  \\
        PruMerge~\cite{shang2024llava} & 18 & 88.0 & 74.4 & 69.7 &62.5 & 57.3 & 81.4 & 66.0  \\
        Matryoshka Multi.~\cite{cai2024matryoshka} & 16 & 90.3 & 81.4 & 80.1 & 70.2 & 67.9 & 75.7 & 75.8 \\
        Matryoshka Query~\cite{hu2024matryoshka} & 16 & 89.3 & 81.4 & 79.2 & 70.6 & 64.7 & 73.8 & 73.6 \\
\rowcolor{Mycolor1} \name~ (Ours) (\texttt{LLaVA-1.5-7B}) & 7.06 & \textbf{98.1} & \textbf{89.5} & \textbf{82.7} & \textbf{77.8} & \textbf{78.1} & \textbf{95.3} & \textbf{93.1} \\
\lasthline
\end{tabular}
}
\vspace*{-0.25cm}
\end{table*}

\begin{table}[!ht]
  \small
    \centering
      \caption{Comparison with various token compression methods on TextCaps dataset for image captioning in terms of BLEU-4 (B@4), CIDEr score, METEOR (MET.) and ROUGE-L.}
    \label{tab:appendix-generative-captioning-evaluation}
     \resizebox{0.5\textwidth}{!}{
    \begin{tabular}{|l|c|c|c|c|c|c|c|c|c|}
 \firsthline
        \rowcolor{airforceblue!20}
         Method & Tokens & B@4 & CIDEr & MET. & ROUGE \\
         \hline
LLAVA-1.5~\cite{liu2024improved} &  576  & 27.1 & 	90.4 & 	21.9 & 	46.2  \\
\hline
PruMerge~\cite{shang2024llava} & $\approx$32 & 17.6 &	62.8 & 	17.0 & 	39.7 \\
Matryoshka Multi.~\cite{cai2024matryoshka}   & 36 & 25.1 & 	94.8 & 	23.0 & 	46.3      \\ 
 Matryoshka Query~\cite{hu2024matryoshka}   & 36  & 21.0 & 	70.0 & 	19.9 & 	42.6    \\
 \rowcolor{Mycolor1} \name~ (Ours)  &  32 & 26.5 & 	90.6 &	22.4  &	46.1\\
\hline
 Matryoshka Query~\cite{hu2024matryoshka}  & 16 & 20.1 & 	62.5  &	19.3 & 	41.7  \\
 \rowcolor{Mycolor1} \name~ (Ours)  &  16 & 26.4 &	90.5 & 	22.5 & 	46.3\\
 \hline
 Matryoshka Query~\cite{hu2024matryoshka}       & 4 & 15.2 & 	42.0 & 	16.5 &	37.4     \\
\rowcolor{Mycolor1} \name~ (Ours)  &  4 & 25.4 & 	86.1 & 	22.0 	&45.7  \\


        \lasthline
    \end{tabular}
    }
    \vspace*{-0.2cm}
\end{table}

\paragraph{Additional discriminative comparisons with other token-summarization approaches.} We note that our approach is the only one that compresses the vision tokens into a representation suitable both for generative and discriminative tasks, requiring no additional forward passes. However, herein, for completeness, we evaluate on our suite of discriminative tasks the current state-of-the-art token compression models that offered pretrained models. This is achieved by following the zero-shot setup described in Paper Section 3.1 
and~\cite{jiang2024e5}. Unsurprisingly, as the results from Table~\ref{tab:appendix-zero-shot-retrieval} and~\ref{tab:appendix-sota_eval_sc} show, our approach significantly surpasses the other methods we compare with.

\paragraph{Additional zero-shot image captioning evaluations:} In addition to the evaluation from the main manuscript, herein, we evaluate our approach for zero-shot captioning on TextCaps~\cite{sidorov2020textcaps}, a dataset for image captioning with reading comprehension. As the results from Table~\ref{tab:appendix-generative-captioning-evaluation} show, we generally match the full-tokens LLaVA's model performance. Importantly, our results remain stable as the number of compressed tokens decreases.

\paragraph{Full-attention vs causal:} Vicuna, and hence LLaVA, much like the rest of the generative LLMs, employs causal attention masking in order to restrict the past states from attending the future ones.  While necessary for autoregressive modeling, it's unclear why it would be for vision token compression too, as there is no preferential direction for image processing. Hence, herein, we explore the effect of changing the attention pattern from causal to bidirectional (\ie full) attention for the compression forward pass, while keeping it causal for the subsequent answer generation ones. In this instance, the stage (\ie compression vs generative) specific LoRAs also take the role of adjusting the attention pattern and information flow. Analyzing the results from Table~\ref{tab:ablation_attention_pattern} we can observe performs gains for discriminative tasks and degradation for generative ones. This suggests that a direct finetuning under a different attention pattern is suboptimal, likely requiring a pre-alignment step. Moreso, the LoRA adapters may limit the ability of the model in shiftting its attention pattern.

\begin{table}[!htbp]
  \centering
  \small
  \caption{Compression with Bidirectional vs Causal attention for generation (MMB, MME, TextVQA) and retrieval (Flickr30K, MS-COCO).}\label{tab:ablation_attention_pattern}
  \resizebox{0.5\textwidth}{!}{
    \begin{tabular}{|l|c|c| c|c|c|c|c|c|c|} 
    \firsthline
     \rowcolor{airforceblue!20}
       &  &   &  & \multicolumn{2}{c|}{ Flickr30K} & \multicolumn{2}{c|}{ MS-COCO} \\ 
     \cline{5-6}\cline{7-8}
      \rowcolor{airforceblue!20}
    \multirow{ -2}{*}{Method} & \multirow{ -2}{*}{MMB}  & \multirow{ -2}{*}{MME} & \multirow{ -2}{*}{TextVQA} & T2I & I2T & T2I & I2T   \\ 
    \hline
Bidirectional & 60.2 & 1310.1 & 48.4 & 83.6 & 94.8 & 57.9 & 72.2  \\
Causal  & 64.4 & 1470.0 & 54.2 & 83.8 & 94.4 & 56.8 & 70.2 \\
\lasthline
\end{tabular}
}
\end{table}

\paragraph{Finetuning checkpoint choice:} The natural starting point for our approach is the LLaVA model itself. However, for completeness, we also try to directly finetune from the Vicuna LLM itself. As the results from Table~\ref{tab:ablation_starting_checkpoint} show, starting from a model already optimized for vision-language understanding results in a notable performance boost. To compensate for this, likely, a longer training scheduler is needed and potentially a full model finetuning, as in LLaVA.

\begin{table}[!htbp]
  \centering
  \small
  \caption{Impact of the pre-trained checkpoint for generation (MMB, MME, TextVQA) and retrieval (Flickr30K, MS-COCO).}\label{tab:ablation_starting_checkpoint}
  \resizebox{0.5\textwidth}{!}{
    \begin{tabular}{|l|c|c| c|c|c|c|c|c|c|} 
    \firsthline
     \rowcolor{airforceblue!20}
       &  &   &  & \multicolumn{2}{c|}{ Flickr30K} & \multicolumn{2}{c|}{ MS-COCO} \\ 
     \cline{5-6}\cline{7-8}
      \rowcolor{airforceblue!20}
    \multirow{ -2}{*}{Method} & \multirow{ -2}{*}{MMB}  & \multirow{ -2}{*}{MME} & \multirow{ -2}{*}{TextVQA} & T2I & I2T & T2I & I2T   \\ 
    \hline
Vicuna & 60.3 & 1296.3 & 48.2 & 81.2 & 92.5 & 54.3 & 67.4  \\
LLaVA  & 64.4 & 1470.0 & 54.2 & 83.8 & 94.4 & 56.8 & 70.2 \\
\lasthline
\end{tabular}
}
\end{table}

\end{document}